# Nearest Neighbor-based Rendezvous for Sparsely Connected Mobile Agents

Ahmad A. Masoud
King Fahd University of Petroleum and Minerals (KFUPM), P.O. Box 287, Dhahran, 31261, Saudi Arabia, e-mail: masoud@kfupm.edu.sa

*Abstract* - In this paper a convergent, nearest-neighbor, control protocol is suggested for agents with nontrivial dynamics. The protocol guarantees convergence to a common point in space even if each agent is restricted to communicate with a single nearest neighbor. The neighbor, however, is required to lie outside an arbitrarily small priority zone surrounding the agent. The control protocol consists of two layers interconnected in a provably-correct manner. The first layer provides the guidance signal to a rendezvous point assuming that the agents have first order dynamics. The other layer converts in a decentralized manner the guidance signal to a control signal that suits realistic agents such as UGVs, UAVs and holonomic agents with second order dynamics.

Rendevous control, Decentralized agents, nearest neighbor protocol, Sparse communication graphs

List of Symbols:

- $L$: number of neighbors an agent observe in order to factor their positions in the direction along which it is going to move
- $N$: number of agents in the group
- $M$: dimensionality of the space in which agents are operating
- $X$: is a vector representing the space in which all agents are operating ($X \in R^M$)
- $X_i$: is a vector representing the center of the i'th agent $X_i$ is a point in $X$
- $\dot{X}_i, \ddot{X}_i$: the velocity and acceleration respectively of the center of agent i
- $d_{i,j}$: is the distance between the center of the i'th and j'th agents
- $\mathbf{D}$: is a distance matrix whose entries are $d_{i,j}$.
- $\beta_i(X)$: is a spherical region in $X$ whose center is $X_i$
- $\epsilon$: is the radius of $\beta_i(X)$
- $Xo_{i,j}$: the centers of all the agents arranged in an ascending order based on their distance from the i'th agent such that $|Xo_{i,j}-X_i| \le |Xo_{i,k}-X_i|$ if $j \le k$
- $Xp_{i,j}$: the centers of all the agents priority arranged based on their distance from the i'th agent
- $S_i(X)$: is a spherical region in $X$ whose center is $X_i$ and radius $dx_i$ equal to the distance between the center of the i'th agent and the furthest agent from it
- $\sigma_i(X)$: is a spherical region in $X$ with center $X_i$ such that $S_i(X) \supset \sigma_i(X)$ where the region $(\sigma_i(X)-\beta_i(X))$ contains one agent only
- $dm_i$: the distance between the i'th agent and the agent in $\sigma_i$
- $Xc$: is the point in $X$ at which all the agents will rendezvous
- $d_{xm}$: is the largest distance among the set of distances corresponding to the closest priority agent to each member of the group
- $Uc_i$: the rendezvous protocol for a single integrator agent (guidance protocol)
- $n_i$: unit vector orthogonal to $Uc_i$
- $Ud_i$: the rendezvous control protocol for a dynamical agent
- $Tc$: convergence time
- $H$: is a hyper vector that contains the centers of all the agents in the group ($H \in R^{M \cdot N}$)
- $\dot{H}$: the time derivative of $H$
- $V(H)$: a common Lyapunov function
- $\lambda$: is a vector that contains the variables describing motion of a nonholonomic agent in its local frame
- $P$: a vector describing the pose of a UGV in two dimensional space. It consists of the position and orientation of the robot $P=[x\ y\ \theta]^T$

# I. Introduction

Polyphase systems have applications in many important areas such as decision making, planning, computer networks and robotics. They consist of subsystems and a logical rule that governs switching among them [1]. In robotics, such systems provide, among other things, the backbone of the process needed to recall to a base-station a group of mobile agents that are exploring a GPS-denied and RF-challenged environment. A polyphase system of such a type drive each member of the group to those closest to it in an attempt to make all agents rendevous at the same location. The control action projected by such a class of systems is called nearest neighbor rendezvous control [2,3].

Nearest neighbor, velocity consensus was first examined by Vicsek et al. [4]. The work modeled the ability of a flock of birds to converge to the same heading through local averaging of the headings of a member's neighbors. An analysis of this behavior was carried-out in [5] by Jadbabaie et al. Cucker & Smale [6] suggested a distance tunable model for velocity consensus where each member of the flock interacts with all other members. They provided conditions for convergence to the same heading that depend only on the initial state of the flock. Convergence to a synchronous velocity state through asynchronous interaction was also observed in [7]. A decentralized, nearest-neighbor, multi-agent controller was designed to de-conflict the use of shared, cluttered space. It was noticed in some of the simulation results that synchronous platoons spontaneously formed in order to avoid possible conflict among agent groups with independent goals.

Design and analysis of protocols that guarantee convergence of agents with single integrator dynamics to a common location in space [8-13] is a major focus of attention. Besides convergence, work in the area is concerned with issues such as: ability to converge in the presence of noise [14] and ability to enforce practical constraints on the process [15]. Despite their simplicity, efficiency and practicality, nearest neighbor protocols may not be able to guarantee convergence. This shortcoming is most visible when the protocol is required to make a group of agents rendevous at the same location in space. The matter is more complicated by the fact that real-life agents can have involved dynamics. Establishing consensus among agents with nontrivial dynamics is challenging. Some of the challenges are: accommodating a variety of practical agents' dynamics, coping with communication delays and enabling the agents to operate in the presence of external drift forces. There is also the issue of maintaining an acceptable transient behavior during the effort to establish consensus. Even if a protocol is convergent and possess acceptable behavior for the single integrator case, the presence of involved dynamics may make these properties impossible to attain [16-19]. Deriving provably-correct, rendezvous control protocols for such a case seem to focus on two types of dynamics. These dynamics are: second order dynamics and nonholonomic dynamics describing the motion of practical mobile agents. It ought to be mentioned that intermittent communication causes another serious problem besides deadlock. To control agents with non-trivial dynamics, exchange of both position and velocity (intentions) information is needed. Obtaining reliable, servo-grade, velocity estimates under intermittent communication/sensing conditions is not possible.

Almost all the work on rendezvous for nonholonomic mobile robots seems specific for the unicycle model [20-23]. The unicycle model is an oversimplification used in modeling nonholonomic agents. Most importantly, it only describes the motion of an agent's body and does not provide the control signal (e.g. speed of the agent's wheels) needed to actualize this motion. A considerable amount of work was done to derive and examine protocols for agents with double integrator dynamics [24]. In [25] the effect of delays on these systems was examined using a directed communication tree. An upper bound on the delays below which the system remains stable was derived. The work assumes that each agent has the position and velocity estimates of the other members it is interacting with. [26] examined the effect of external disturbances on the ability of double integrator agents to reach consensus. It showed that the error can be bounded in a leader-follower setting. However, an estimate of the leader's velocity is required by the followers. [27] tackled saturation in the actuators of double integrator agents. An auxiliary system that defines appropriate intermediate reference trajectories is used to reduce the design of the consensus protocol to that of the ideal situation without input saturations and in the full state feedback case.

This work has two parts. The first part offers a variant of the traditional, nearest neighbor consensus protocol. The modified protocol guarantees convergence of a group of agents with single integrator dynamics to a common rendezvous point in space. It only requires each agent to be able to communicate with at least one other neighbor. It is enough that this neighbor be the one closest to the agent provided that it lies outside an arbitrarily small priority zone surrounding the agent concerned. The second part of the paper has to do with guaranteeing stability when the protocol is used by a group of dynamical agents. The suggested procedure for converting the guidance signal from the rendezvous protocol to a rendezvous control signal does not require exchange of velocity information among the participants (i.e. exchange intentions). Along with the guidance signal each agent uses its own velocity information for generating the control signal. The procedures have several advantages. For example, they demonstrate significant robustness in the presence of communication delays, actuator saturation and external disturbances. Moreover, they provide well-behaved transient response and control signals. The generation of the control protocol is based on a series of methods suggested by this author. The methods convert the guidance planning signal from a harmonic potential [28,29] to a control signal for holonomic systems with second order dynamics [30], nonholonomic mobile robots [31] and a large class of autonomous agents and UAVs [32]. It is proven in this paper that these techniques, which were originally designed for a single agent, are fully capable of functioning as control protocols in a multi-agent environment. The close relation between harmonic potential and the consensus problem is the main motivation for examining the use of these techniques. The value of a harmonic potential at a point is arrived at iteratively as the average value of its immediate neighbors.

## II. The rendezvous protocol

Nearest neighbor-based consensus protocols are important and practical. Whether the information exchange among the agents is based on sensing or communication, nearest neighbor-based action always has the best chance succeeding. Existing protocols, however, do not guarantee convergence to a common location under practical conditions which the operator can *a priori* know and set. An example of this is the number of neighbors each agent has to communicate with in order to guarantee convergence to a meeting point. While it is known that agreement among the group members may be established if each agent is communicating with sufficient neighbors, this number cannot be exactly known prior to operation. This is demonstrated in Figure-1. The effect of increasing the number of neighbors (L) on the ability of 50, single-integrator agents to agree on a meeting point is tested. As can be seen, each agent has to communicate with at least 20% of the group (L=10) in order for an agreement to be reached. An RF-challenged environment will most probably not support such a large communication overhead.

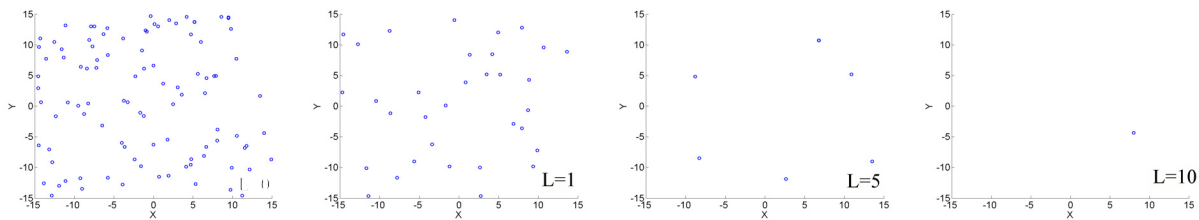

Figure-1: nearest neighbor consensus protocol does not guarantee convergence

It is most likely that the main cause of the convergence problem in the traditional nearest-neighbor consensus protocol has to do with the manner in which the effort to establish agreement is distributed. It does not make sense for an agent to spend effort establishing agreement with other agents who are already in agreement with it. Such agents may be considered as one agent with multiplicity. Agents with large, but manageable, deviations from the actor agent should have high priority. Others whose state is close to the agent concerned should have low priority as far as dispensing the consensus effort is concerned.

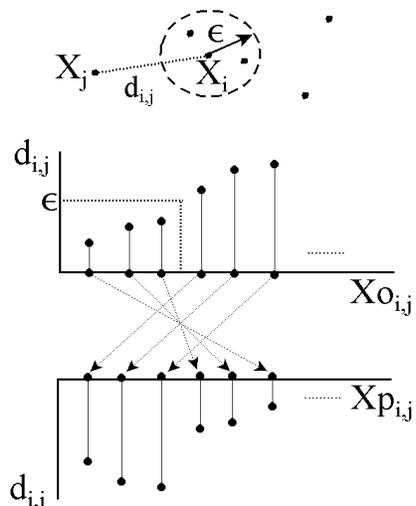

Figure-2: priority buffer arrangement

Factoring-in priority in a manner that provides a practical solution to the convergence problem may be done as follows: consider an M-dimensional sphere (M is the dimensionality of the space in which the agents are operating) of radius $\epsilon$ ($\beta_i(X)$) that is centered around the position of the i'th agent ($X_i$)

$$\beta_i(X) = \{X : |X - X_i| \leq \varepsilon\}. \qquad (1)$$

If the j'th agent (i≠j) is in $\beta_i$ ( $X_j \in \beta_i$ ), this agent is considered as a low priority agent; otherwise, it is a high priority agent. Let $d_{ij}$ be the distance between the i'th and the j'th agents ($d_{i,j} = |X_i - X_j|$). Also let $Xo_{i,j}$ be a buffer containing the locations of the agents ordered in an ascending manner based on their distance from $X_i$ ($d_{i,j-1} \leq d_{i,j} \leq d_{i,j+1}$, i=1,..,N, j=1,...N, i≠j). Existing consensus protocols that use the L closest neighbors generate the velocity vector of the i'th agent as

$$\dot{X}_i = Uc_i = \sum_{j=1}^{L} a_{i,j}(Xo_{i,j} - X_i) \quad (2)$$

where $a_{ij}$ are positive constants.

The modified protocol works as follows: first, the protocol priority orders (figure-2) the agents relative to the i'th agent ($Xp_{i,j}$). $Xp_{i,j}$ is constructed as follows

$$Xp_{i,j} = Xo_{i,j+Lo} \quad j = 1,..,N - Lo$$
$$Xp_{i,j+Lo-1} = Xo_{i,L0-j+1} \quad j = 1,...,Lo \quad (3)$$

where $d_{i,Lo} < \epsilon$, N is the total number of agents and Lo is the number of agents inside $\beta_i$. In a similar manner to the normal protocol, the velocity of the i'th agent is constructed as

$$\dot{X}_i = Uc_i = \sum_{j=1}^{L} a_{i,j}(Xp_{i,j} - X_i) \quad (4)$$

where L is the number of agents the modified protocol is using to generate the velocity of the i'th agent. It is important to notice that the manner in which the protocol is modified neither affects its scalability or its ability to maintain a decentralized mode of operation. Also, in practical implementations, signal strength & directivity may be used as representatives of a neighboring agent relative distance and orientation.

## III. Protocol Analysis

This section examines the behavior of the suggested protocol in terms of ability to guarantee convergence to a rendezvous point, the effect of the priority zone on the rate of convergence and the communication burden during the agents' attempt to reach a rendevous point.

III.1- Convergence:
The modified protocol is convergent provided that the i'th agent maintains connection with at least one agent outside $\beta_i$. Three distinct hyper-spheres (figure-3) whose center is $X_i$ are used in the proofs: The previously defined priority zone, $\beta_i$, in which $X_i$ is guaranteed to communicate with all agents in that zone. A hyper sphere, $S_i$, containing all the agents of the group is defined. The center of $S_i$ is $X_i$. Its radius, $dx_i$, is selected as the distance between $X_i$ and the agent furthest from it. In this section, it is assumed that $X_j$ is the agent furthest from $X_i$,

$$S_i(X) = \{X : |X - X_i| \leq dx_i\} \quad X_i \in S_i(X) \, \forall \, i \quad (5)$$

The last sphere with $X_i$ as a center is $\sigma_i$ ($\sigma_i \subset S_i(X)$). This is the largest hollow sphere ($\beta_i$ is not included) containing only one agent which is the agent closest to $X_i$ that does not belong to $\beta_i$. In this section, this agent is referred to as $X_k$,

$$X_k \cap \sigma_i = X_k, \quad X_i \cap \beta_i = \emptyset \quad i \neq k \quad (6)$$

It ought to be noticed that by construction, for the case of L=1, the rendezvous protocol will only operate on agents with non-zero distance. Therefore an implicit assumption in the proof is that $X_i \neq X_j$ for any i & j.

Proposition-1: The distance, $dm_i$

$$dm_i = |X_k - X_i| \quad (7)$$

is always decreasing, where $dm_i$ is the distance between agent $X_i$ and the agent closest to it ($X_k$) that lies in $\sigma_i$.

Proof: With no loss of generality, the convergence proof is carried-out assuming unity coefficients $a_{i,j}$. The value of these coefficients influence only the rate of convergence, not convergence itself.

There are two possibilities, either the agent closest to $X_k$ ($X_l$) is in $\beta_i$ or it is outside $\sigma_i \cup \beta_i$. If $X_l \in \beta_i$, then

$$\frac{d(dm_i)}{dt} = -dm_i - dm_k \cdot \cos(\theta_k) \quad (8)$$

where $\quad -\frac{\pi}{2} < \theta_k < \frac{\pi}{2} \quad \& \quad dm_k = |X_l - X_k|.$

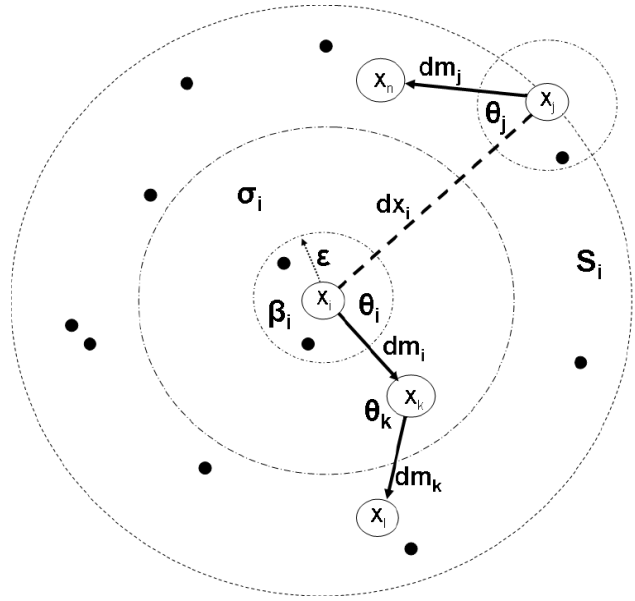

Figure-3: distances relative to agent i

In other words:
$$\cos(\theta_k) > 0$$
and
$$\frac{d(dm_i)}{dt} < 0. \qquad (9)$$

In the second case ($X_k \in S_i - (\beta_i \cup \sigma_i)$), we have
$$dm_k < dm_i \qquad (10)$$
otherwise $X_k$ will be closer to an agent in $\beta_i$. Therefore, regardless of the value of $\theta_k$ equation-9 will hold making the derivative of $dm_i$ strictly negative.

Proposition-2: The modified protocol is globally asymptotically convergent, i.e.
$$\lim_{t \to \infty} X_i = X_c \qquad i=1,...,N \qquad (11)$$
N is the number of agents in the group, $X_c$ is the rendezvous point.

Proof: The proof is carried-out for L=1 connectivity. This proof subsumes the one for L>1 connectivity. The proof is based on showing that the distance, $dx_i$, from an agent i to the agent furthest from it, agent j, will shrink to zero, i.e.
$$\lim_{t \to \infty} dx_i = 0 \qquad i=1,...,N \qquad (12)$$
where
$$dx_i = \max_m |X_m - X_i| \qquad m=1,...,N \qquad (13)$$

The time derivative of $dx_i$ may be written as:
$$\frac{d(dx_i)}{dt} = F_i + F_j \qquad (14)$$

where $F_i$ is the dot product between the velocity vector of agent i ($\dot{X}_i$) and the unit vector from $X_i$ pointing towards $X_j$. $F_j$ is the dot product between the velocity vector of agent j ($\dot{X}_j$) and the unit vector from $X_j$ pointing towards $X_n$, where $X_n$ is the agent closet to $X_j$ outside $\beta j$

$$\frac{d(dx_i)}{dt} = -\frac{(X_i - X_j)^T}{|X_i - X_j|}(\dot{X}_j - \dot{X}_i) \qquad (15)$$

The derivative may be written as:
$$\frac{d(dx_i)}{dt} = -(dm_i \cdot \cos(\theta_i) + dm_j \cdot \cos(\theta_j)) \qquad (16)$$

$dm_j$ is the distance between agent $X_j$ and the agent closest to it ($X_n$) that lies in $\sigma_j$, $\theta_i$ is the angle between lines $dm_i$ and $dx_i$ and $\theta_j$ is the angle between lines $dm_j$ and $dx_i$.

Let's examine the derivative of $dx_i$ in the two zones: $S_i$-$\beta_i$ and $\beta_i$. As shown in proposition-1, in the zone $S_i$-$\beta_i$, $dm_i$ is always decreasing. On-the-other-hand, $\theta_j$ is restricted to lie between

$$-\frac{\pi}{2} < -\cos^{-1}(\frac{\varepsilon}{2 \cdot dx_i}) \leq \theta_j \leq \cos^{-1}(\frac{\varepsilon}{2 \cdot dx_i}) < \frac{\pi}{2}$$

and
$$dm_j \geq \varepsilon \qquad (17)$$

Therefore the time derivative of $dx_i$ will be negative

$$\frac{d(dx_i)}{dt} < 0 \qquad X_j \in S_i - \beta_i \qquad (18)$$

which will guarantee that all agents will converge to $\beta_i$.

Once $X_j$ enter $\beta_i$, the control law moves agent i towards the agent that is furthest from it (agent j). This makes

$$\frac{d(dx_i)}{dt} = -|X_j - X_i| \qquad (19)$$

guaranteeing that
$$\lim_{t \to \infty} dx_i = 0 \qquad i=1,...,N \qquad (20)$$

The fact that all agents will converge to agent $X_i$ for any i can only hold if all agents converge to the same point $X_c$,

$$\lim_{t \to \infty} X_i = X_c \qquad i=1,..N \qquad (21)$$

III.2- Communication range limits

The communication limit on the agents is an important factor in determining the practicality of a rendezvous process. In this regard, the suggested protocol has nonastringent requirements. As shown above, the protocol is guaranteed to converge even if each agent is restricted to communicate with its nearest neighbor outside $\beta$. If an agent cannot communicate with the closest agent, this agent is isolated and cannot participate in the rendezvous effort to begin with. However, the communication limits may still be assessed by examining the behavior of the maximum of the distances connecting each agent in the group to its closest neighbor outside the priority zone corresponding to that agent ($d_{xm}$)

$$d_{xm} = \max_i(\min_j d_{i,j} / d_{i,j} > \varepsilon) \qquad i=1,...,N \qquad (22)$$

where $d_{i,j}$'s are entries in the distance matrix **D**, the i'th row has the distances from agent i to all other agents in the group.

A large increase in $d_{xm}$ during operation may jeopardize the ability of the agents to communicate. As shown below, the protocol can inhibit the growth of $d_{xm}$, hence prevent the communication burden from increasing during the agents' effort to reach a meeting point.

$$\mathbf{D} = \begin{bmatrix} 0 & d_{12} & d_{13} & . & d_{1N} \\ d_{21} & 0 & d_{23} & . & d_{2N} \\ d_{31} & d_{32} & 0 & . & d_{3N} \\ . & . & d_{ij} & . & . \\ d_{N1} & d_{2N} & d_{3N} & . & 0 \end{bmatrix} \qquad (23)$$

Proposition-3: if an agent j enters into the priority zone of agent i ($\beta_i$) it will remain inside $\beta_i$.

Proof: The proof follows directly from proposition-1. This may also be deduced from the fact that when $X_j$ has just left $\beta_i$ it becomes the minimum distance agent away from $X_i$ and the protocol will steer it back to $\beta_i$ (figure-4).

Proposition-4: If agent-j lies in the intersection of $\beta_i$ and $\beta_k$ (figure-5)
$$X_j \in \beta_i \cap \beta_k$$
then
$$|X_i - X_k| < 2 \cdot \varepsilon. \qquad (24)$$

Proof: this follows directly from proposition-3. If $X_j$ is inside $\beta_i$ and $\beta_k$ then it will always belong to these two regions. This can only happen if equation-24 holds.

Proposition-5: If $\exists$ an $X_j \in \beta_i \cap \beta_k \ \forall i \neq k$, then $\qquad d_{xm} < 2\varepsilon. \qquad (25)$

Proof: this proposition follows directly from proposition-4.

It ought to be mentioned that attempting to control $d_{xm}$ by increasing the connectivity of the graph maybe ineffective in controlling the growth of this distance. While increasing connectivity will accelerate the rendezvous process, it will not prevent the creation of drifting clusters each forming a closed group with agents that are temporarily communicating with each other. To reduce the probability of such clusters forming, connectivity has to be increased to an unrealistically high value.

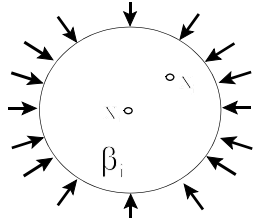

Figure-4: If an agent enters $\beta_i$ it remains in $\beta_i$.

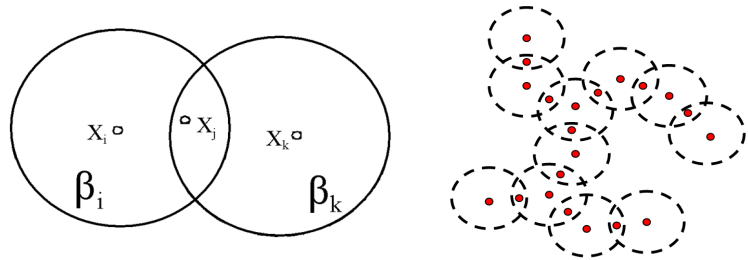

Figure-5: joint sharing of an agent guarantees connectedness

III.3- Convergence Time

In this section an upper bound on the convergence time of the protocol (Tc) is derived. Tc is defined as the time it takes all agents to enter a sphere of radius $\delta$ and stay in it

$$|X_i - X_j| \leq \delta \qquad \forall t \geq Tc, \forall i, j \qquad (26)$$

The bound is a function of two factors. The first is the initial spatial distribution of the agents. The other is the width of the priority zone ($\epsilon$). The $\epsilon$-dependance clearly shows that the bound on Tc quickly decays and settles to a constant value. This means that a large value of $\epsilon$ does not hold any significant advantage in terms of convergence compared to a small one. A large value of $\epsilon$ could cause an issue as far as the communication ability of the agents in $\beta_i$ are concerned.

Consider a spherical region of width $\delta$ around the i'th agent ($X_i$). The following cases may be used to derive an upper bound on Tc:

1- $\epsilon=0$: For this case the protocol reduces to the conventional rendezvous protocol whose convergence is not guaranteed (i.e. Tc$\leq\infty$).
2- $\epsilon \geq dx_i(0)$ : In this case all the agents are inside the priority zone of the i'th agent. Therefore the convergence time is not a function of $\epsilon$ (i.e. Tc=C, where C is a constant).
3- $dx_i > \epsilon > \delta$: In this case Tc is the sum of two parts: The first part is the time it takes all agents to enter the i'th priority zone (T$\epsilon$). The second is a constant $C_1$ representing the time for the agents in the priority zone to all enter the $\delta$-sphere (similar to case two)

$$Tc=T\epsilon+C_1 \qquad (27)$$

The derivation of the upper bound is carried-out for L=1 connectivity. Since it is well-known that increasing connectivity among agents will expedite convergence this case will provide an upper bound on Tc.

The dynamic equation governing the distance of the n'th agent from the i'th agent, $d_{i,n}$, (assuming that the n'th agent lie in $\sigma_i$) is

$$\dot{d}_{i,n} = -a \cdot d_{i,n} \qquad (28)$$

where a is a positive constant. This leads to the solution

$$d_{i,n}(t) = \exp(-a \cdot t)d_{in}(0) \qquad (29)$$

The time needed for $d_{i,n}$ to drop below $\epsilon$ (i.e. the n'th agent to enter $\beta_i$) is

$$-\frac{1}{a}\ln(\frac{\varepsilon}{d_{i,n}(0)}) \quad . \tag{30}$$

The time needed for all agents to enter the i'th priority zone may be bounded as:

$$\begin{aligned} T\varepsilon &\leq \sum_{n=1}^{K} -\frac{1}{a}\ln(\frac{\varepsilon}{d_{i,n}(0)}) \\ &\leq \sum_{n=1}^{K} -\frac{1}{a}\ln(\frac{\varepsilon}{dx_i(0)}) \\ &= -\frac{K}{a}\ln(\frac{\varepsilon}{dx_i(0)}) \end{aligned} \tag{31}$$

where K is a positive integer. The convergence time of the rendezvous protocol may be bounded as:

$$Tc \leq -\frac{K}{a}\ln(\frac{\varepsilon}{dx_i(0)}) + C_1. \tag{32}$$

While K, $C_1$ and $dx_i(0)$ are *a priori* unknown, the value of the above expression lies in the pattern the bound on Tc exhibits as a function of $\epsilon$. The bound shoots to infinity when $\epsilon$ is zero and rapidly decays to a constant value as $\epsilon$ increases. This behavior is demonstrated by simulation in section-V (figure-12).

## IV. Rendevous for dynamical agents

The agents that may need to implement the above protocol are most likely to have involved dynamics. The interaction of the agents' dynamics with the rendezvous protocol may result in a loss of ability of the agents to reach an agreement. Converting the protocol to a control protocol is a challenging task. The challenge is to achieve system stabilization as well as compliance with the guidance signal from the protocol using local information and actions. In other words, each agent must use its own state to synthesize a successful, self-control action. A series of work by this author on the above subject proves to be promising. The control schemes were designed for a single agent to effectively suppress motion in the space orthogonal to the guidance vector. This paper proves that this approach to design makes these controllers valid control protocols that may be successfully used in a multi-agent environment.

IV.1 The Double Integrator Rendezvous Control Protocol
The simplest holonomic agent with second order dynamics has the form

$$\ddot{X}_i = U_i \quad i=1,..N \tag{33}$$

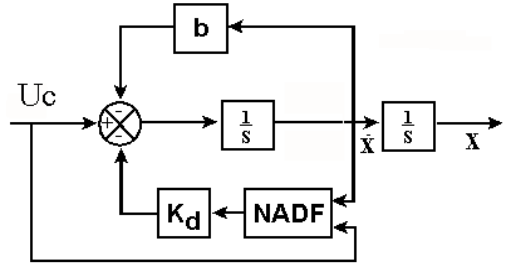

Figure-6: NADF based control protocol

The simplistic way of constructing a control protocol for this case is to augment the guidance protocol with a damping term constructed from the agents' velocities

$$\ddot{X}_i = Uc_i - b\dot{X}_i \quad i=1, .. N \tag{34}$$

If a small linear damping coefficient b is used, the dynamical interactions among the agents may prevent convergence. If an excessively large b is used, motion will be severely impeded and the rate of convergence will be drastically reduced. To solve this problem the concept of nonlinear, anisotropic damping forces (NADF) was suggested in [30]. NADFs (Figure-6) selectively apply high motion impedance ($Un_i$) in the space orthogonal to $Uc_i$

$$Un_i = [(n_i^T \dot{X}_i n_i + (\frac{Uc_i^T}{|Uc_i|} \cdot \dot{X}_i \Phi(Uc_i^T \dot{X}_i)) \frac{Uc_i}{|Uc_i|}] \tag{35}$$

where $n_i$ is a unit vector orthogonal to $Uc_i$ and $\Phi$ is the heaviside function. The control protocol in this case is

$$Ud_i = Uc_i - b\dot{X}_i - K_d Un_i. \tag{36}$$

Excessively high value of $K_d$ may be used without affecting the rate of convergence or quality of the control signal.

Proposition-6: For any $b>0$ and $K_d >0$, the protocol for the second order system in (36) guarantees convergence of the system in (33).
$$\lim_{t \to \infty} X_i = X_C \qquad i=1,..,N \qquad (37)$$

Proof: Define the hyper sate $H=[X_1^T \ X_2^T \ ... \ X_N^T]^T$. Since the protocol in (4) is globally, asymptotically convergent, there exists a dynamical system
$$\dot{H} = G(H) \qquad (38)$$
such that
$$\lim_{t \to \infty} H = H_C$$
where $H_C = [X_C^T \ X_C^T \ ... X_C^T]^T$ and $G=[Uc_1^T \ Uc_2^T \ ... Uc_N^T]^T$. It is important to notice that the components of (38) are stable systems. A converse Lyapunov theory in [33,34] shows that a common Lyapunov function ($V(H)$) does exist for the system in (38),
$$V(H) = \begin{bmatrix} 0 & H = H_C \\ +\text{ve} & \text{otherwise} \end{bmatrix} \qquad (39)$$
and
$$\dot{V}(H) = \begin{bmatrix} 0 & H = H_C \\ -\text{ve} & \text{otherwise} \end{bmatrix} \qquad (40)$$

To prove convergence of the protocol in (36), the following Lyapunov function is constructed
$$V_d(H, \dot{H}) = V(H) + \frac{1}{2}\dot{H}^T\dot{H} \qquad (41)$$

The time derivative of $V_d$ is
$$\begin{aligned} \dot{V}_d &= -G^T(H)\dot{H} + \dot{H}^T\ddot{H} \\ &= -G^T(H)\dot{H} + \dot{H}^T(G(H) - b \cdot \dot{H} - k_d[Un_1 .... Un_N]^T) \\ &= -b \cdot \dot{H}^T\dot{H} - k_d \sum_{i=1}^{N} \dot{X}_i^T Un_i \end{aligned} \qquad (42)$$

notice that
$$\dot{X}_i^T Un_i = (\dot{X}_i^T n_i)^T (\dot{X}_i^T n_i) + \frac{1}{|Uc_i|^2}(\dot{X}_i^T Uc_i)^T (\dot{X}_i^T Uc_i) \qquad (43)$$

As can be seen the time derivative of $V_d$ is negative semidefinite with a zero set Z
$$Z = \{H, \dot{H} = 0\}. \qquad (44)$$
The minimum invariant set (E) of the system
$$\ddot{H} = G(H) - b \cdot \dot{H} - k_d \begin{bmatrix} Un_1 \\ . \\ . \\ Un_N \end{bmatrix} \qquad (45)$$

is computed using $G(H)=0$ and yields $E=\{H_C\}$. By the LaSalle invariance principle [35], the system will converge to the set $E \cap Z = H_C$. In other words
$$\lim_{t \to \infty} X_i = X_C. \qquad i=1,..,N \qquad (46)$$

IV.2 Rendezvous for UGVs

In [31] a method is suggested for converting $Uc_i$ into a control signal for a UGV whose system equation may be written as
$$\dot{P} = F(P)\lambda \qquad (47)$$
$$\lambda = Q(U)$$
where P is the posture of the UGV, $\lambda$ is the velocity in the local coordinates and U is the control signal.

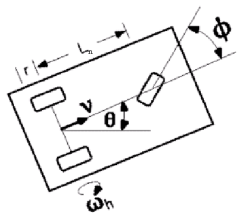

Figure-7: A car-like mobile robot

Many practical UGVs do fit the above system equation including the car-like, front wheel-steered UGV (figure-7) with

system equation and control protocol

$$\begin{bmatrix} \dot{x} \\ \dot{y} \\ \dot{\theta} \end{bmatrix} = \begin{bmatrix} \cos(\theta) & 0 \\ \sin(\theta) & 0 \\ 0 & 1 \end{bmatrix} \begin{bmatrix} v \\ \omega \end{bmatrix}, \qquad (48)$$

$$\begin{bmatrix} v \\ \omega \end{bmatrix} = \begin{bmatrix} r \cdot \omega_h \\ r \cdot Ln \cdot \omega_h \cdot \tan(\varphi) \end{bmatrix} \quad \begin{bmatrix} \omega_h \\ \phi \end{bmatrix} = \begin{bmatrix} vc/r \\ \tan^{-1}(\Delta\theta/(Ln \cdot vc/r)) \end{bmatrix},$$

$$vc = K_1 \cdot |Uc_i| \quad \& \quad \Delta\theta = K_2(\arg(Uc_i) - \theta)$$

where $P=[x\ y\ \theta]^t$, $\lambda=[v\ \omega]^t$, $U=[\omega_h\ \phi]^t$, r is the radius of the robot's wheels,, v is the tangential velocity of the robot and $\omega$ is its angular speed, Ln is the normal distance between the center of the front wheel and the line connecting the rear wheels, $\omega_h$ is the angular speed of the rear wheels, and $\phi$ is the steering angle of the front wheel ($\pi/2 > \phi > -\pi/2$).

Proposition-7: For $K_1 > 0$, $K_2 > 0$ and $K_2 \gg K_1$, the control protocol for the non-holonomic system in (47)

$$Ud_i = Q^{-1}(\lambda c_i) \quad i=1,...N \qquad (49)$$

where

$$\lambda c_i = \begin{bmatrix} K_1 \cdot |Uc_i| \\ K_2 \cdot (\arg(Uc_i) - \theta_i) \end{bmatrix} \qquad (50)$$

will guarantee that

$$\lim_{t \to \infty} \begin{bmatrix} x_i \\ y_i \end{bmatrix} = \begin{bmatrix} xc \\ yc \end{bmatrix} \quad i=1,..,N \qquad (51)$$

Proof: substituting the control system protocol in (49,50) in the robot's system equation (48) yields the two equations

$$\begin{bmatrix} \dot{x}_i \\ \dot{y}_i \end{bmatrix} = K_1 \cdot \begin{bmatrix} |Uc_i| \cdot \cos(\theta_i) \\ |Uc_i| \cdot \sin(\theta_i) \end{bmatrix} \qquad (52)$$

$$\dot{\theta}_i + K_2 \cdot \theta = s(t) \qquad (53)$$

where $s(t)=K_2 \cdot \arg(Uc_i)$.

The followings may be deduced:
1- equation (53) is independent of equation (52)
2- since $K_2$ is high, the natural response of the nonhomogeneous, linear, first order differential equation in (53) does approximate an impulse. Therefore, the nonhomogeneous solution of the equation, which is equal to the natural response convolved by s(t) may be approximated as: $\theta_i \approx \arg(Uc_i)$.
3- Since $K_2 \gg K_1$, one may approximate (52) as

$$\begin{bmatrix} \dot{x}_i \\ \dot{y}_i \end{bmatrix} = K_1 \cdot \begin{bmatrix} |Uc_i| \cdot \cos(\arg(Uc_i)) \\ |Uc_i| \cdot \sin(\arg(Uc_i)) \end{bmatrix} \qquad (54)$$

Let's use the common Lyapunov function (V(H)) in 10, where $H=[\ [x_1\ y_1]\ ...\ [x_N\ y_N]\ ]^T$. The time derivative of V(H) is

$$\dot{V}(H) = \nabla V(H)^T \dot{H}^T \qquad (55)$$

Substituting (54) in (55), we have

$$\dot{V}(H) = K_1 \cdot \nabla V(H)^T \begin{bmatrix} |Uc_1| \cdot \cos(\arg(Uc_1)) \\ |Uc_1| \cdot \sin(\arg(Uc_1)) \\ \vdots \\ |Uc_N| \cdot \cos(\arg(Uc_N)) \\ |Uc_N| \cdot \sin(\arg(Uc_N)) \end{bmatrix} \qquad (56)$$

Since the original protocol in (49,50) is convergent, the time derivative of V(H) (which is the same as the one in (56) is negative definite. In other words, the control protocol for the nonholonomic robot in (47) is also convergent.

IV.3 Rendezvous for UAVs
In [32] a control structure (figure-8) is suggested for converting the guidance signal in a provably-correct manner to a control protocol that suits a dynamical system of the form

$$\dot{X} = F(\lambda)$$
$$\dot{\lambda} = Q(\lambda, U) \quad (57)$$

where U is the control signal, X is a vector containing the location of the center of mass of the UAV in the world coordinates, $X=[x\ y\ z]^t$, $\lambda$ is the motion vector in the local coordinates of the UAV, $\lambda=[v\ \gamma\ \psi]^t$, $v$ is the tangential speed of the UAV, $\gamma$ and $\psi$ are angles describing its orientation with respect to the world coordinates. This model suits most UAVs. A specific form for equation (57) that describe a fixed-wing (figure-9) aircraft is shown in equation (58),

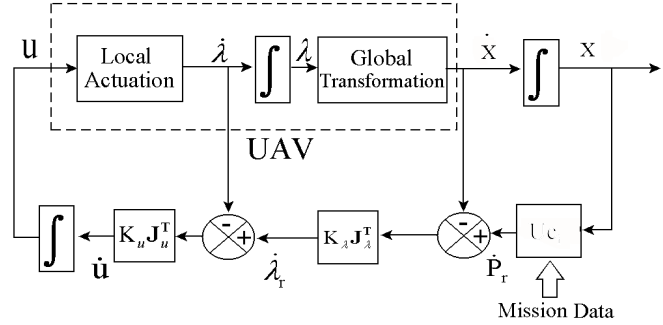

Figure-8: converting guidance into control for a UAV

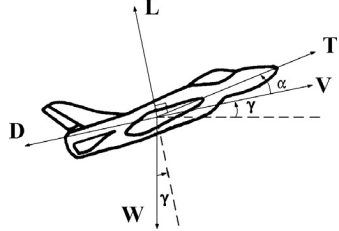

Figure-9: A fixed-wing UAV.

$$\begin{aligned}
\dot{x} &= v \cdot \cos(\gamma)\cos(\psi) \\
\dot{y} &= v \cdot \cos(\gamma)\sin(\psi) \\
\dot{z} &= v \cdot \sin(\gamma) \\
\dot{v} &= \frac{F_T}{m} - g \cdot \sin(\gamma) \\
\dot{\gamma} &= \frac{F_N \cdot \cos(\eta)}{m \cdot v} - g \frac{\cos(\gamma)}{v} \\
\dot{\psi} &= \frac{F_N \cdot \sin(\eta)}{m \cdot v \cdot \cos(\gamma)}.
\end{aligned} \quad (58)$$

where m is the point mass of the UAV, $v$ radial velocity of the UAV, $\gamma$ flight path angle, $\psi$ directional angle, $\eta$ is the banking angle, $F_T$ the resultant force along the velocity vector:

$$F_T = T \cdot \cos(\alpha) - L_d \quad (59)$$

and $F_N$ is the resultant force normal to the velocity vector:

$$F_N = T \cdot \sin(\alpha) + L_f \quad (60)$$

and g is the constant of gravity, T is the thrust from the engine, $L_d$ is the aerodynamic drag, $\alpha$ is the angel of attack, $L_f$ is the aerodynamic lift. The control protocol is:

$$\dot{U} = K_u J_u^T \left[ K_\lambda J_\lambda^T (Uc - F(\lambda)) - Q(\lambda, U) \right] \quad (61)$$

$$U(t) = \int_{t_0}^{t} \dot{U} dt \quad (62)$$

where $J_u = \dfrac{\partial F(\lambda, U)}{\partial U}$, $J_\lambda = \dfrac{\partial F(\lambda)}{\partial \lambda}$, $K_u$ and $K_\lambda$ are positive constants.

Proposition-8: for $K_u>0$ and $K_\lambda>0$ the control law in (61,62) may be used as control protocol by agents whose dynamic equation is described by (57) to guarantee convergence to a common rendezvous point Xc

$$\dot{U}_i = K_u J_{u_i}^T \left[ K_\lambda J_{\lambda_i}^T (Uc_i - F(\lambda_i)) - Q(\lambda_i, U_i) \right] \quad (63)$$

$$U_i(t) = \int_{t_0}^{t} \dot{U}_i \, dt \qquad (64)$$

$$\lim_{t \to \infty} X_i = X_C. \qquad i=1,..,N \qquad (65)$$

Proof: As mentioned earlier, the fact that the protocol in (3) is convergent implies that a common Lyapunov function (V(H)) exists and its time derivative is negative definite for the single integrator system:

$$\dot{X}_i = Uc_i \qquad i=1,..,N \qquad (66)$$

Let the same V(H) be used with agents whose dynamic equation is (57). The time derivative of the V(H) is

$$\dot{V}(H) = \nabla V(H)^T \dot{H} \qquad (67)$$

Since the protocol is convergent for the single integrator case, we have

$$\dot{V}(H) = \nabla V(H)^T G(H) < 0 \qquad (68)$$

Define $\Omega$ as the region containing all possible values of G(H) (figure-10). Since it was proven in [32] that if $K_u > 0$ and $K_\lambda > 0$, we have

$$\lim_{t \to \infty} \dot{X}_i = Uc_i \qquad i=1,..N \qquad (69)$$

$\Omega$ is an attractor set for $\dot{H}$

$$\dot{H} \in \Omega \qquad \text{For } t > T. \qquad (70)$$

This in turn implies that

$$\dot{V}(H) = \nabla V(H)^T \dot{H} < 0 \qquad (71)$$

and the agents whose behavior is governed by (57) will converge to $X_C$.

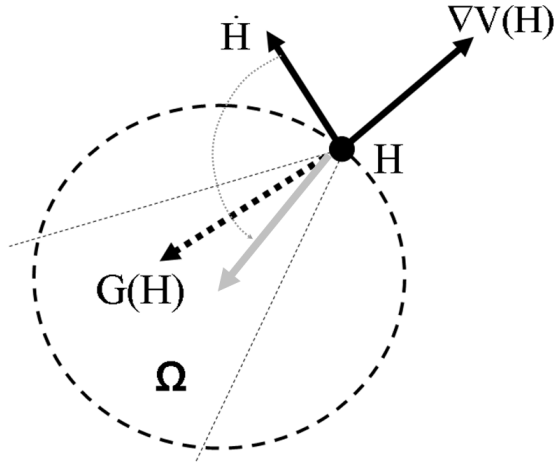

Figure-10: The time derivative of H will converge to $\Omega$.

## V. Simulation Results

In figure-11 the modified rendezvous protocol is tested for 2400 agents with a uniformly-distributed, random initial configuration. Each agent communicates only with one priority nearest-neighbor in its attempt to rendezvous with the others. The radius of the low priority region ($\epsilon$) is arbitrarily set to 1. As can be seen, the agents converged to a point that is close to the average of the initial configurations.

It is well-known that the more neighbors an agent communicate with (i.e. the more connected the communication graph is) the faster convergence will be. However, the effect of $\epsilon$ on convergence need to be examined. The value of $\epsilon$ is varied from zero to a high value and convergence time is measured. The simulation is carried-out for 100 agents each communicate with the closest 5 priority neighbors (figure-12). All other cases showed a behavior similar to the one obtained for this case which is in conformity with (32). When $\epsilon$ is set to zero, i.e. the protocol reduces to the original nearest neighbor algorithm, no convergence took place. It is observed that convergence time rapidly drops as $\epsilon$ increases and settles to a constant value. Small values of $\epsilon$ exceeding .02 do not offer any significant improvement as far as the convergence rate is considered.

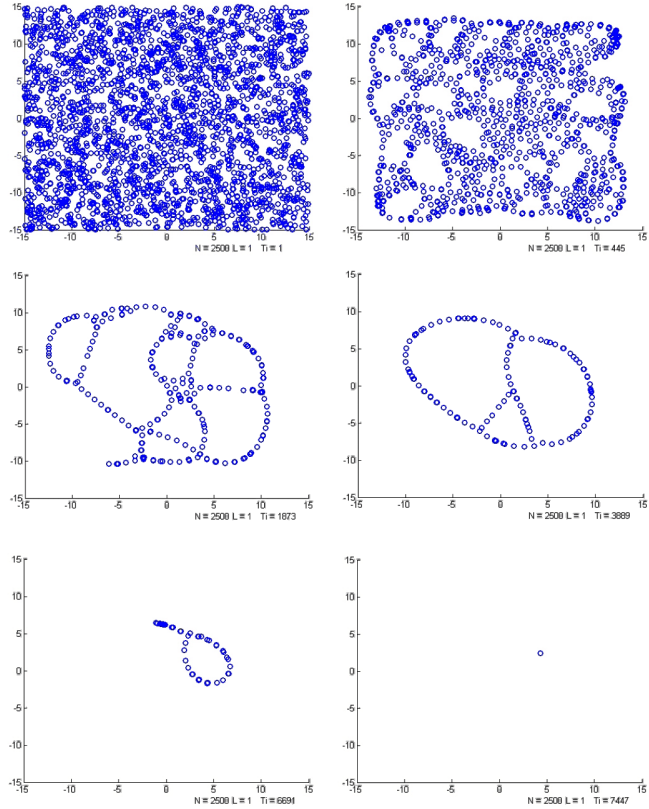

Figure-11: Modified protocol guarantees convergence, L=1.

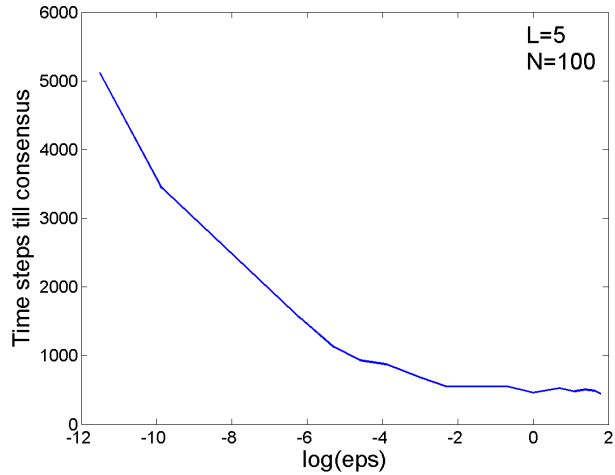

Figure-12: Time to converge versus log($\epsilon$)

In table-1, the effect of $\epsilon$ on $d_{xn}$ is tested for 200 agents initially located in a 30×30 rectangular region and distributed in space using a uniform PDF. The initial $d_{xn}$, maximum $d_{xn}$ and time (Tc) to rendezvous (in time steps) are recorded. The average distance traveled by the agents until consensus is achieved (dL) along with the most distance traveled minus the least distance traveled (Δ) are also recorded. As can be seen, there is a considerable growth in $d_{xn}$ for low values of $\epsilon$. At $\epsilon$=3, condition-25 is satisfied. This restricted the maximum value of $d_{xn}$ for $\epsilon \geq 3$ to less than $2\epsilon$. The changes in $\epsilon$ has minimal effect on the time to reach consensus. Figures 13 & 14 show $d_{xn}$ versus time for $\epsilon$=1 and $\epsilon$=3 respectively. The ability to control the growth of $d_{xn}$ is obvious. Increasing connectivity to control $d_{xm}$ is investigated in table-2. Although the rate of convergence significantly improved, L had practically no effect on $d_{xm}$ until it was set to an unrealistically high value.

| $\epsilon$ | $d_{xn}$ initial | $d_{xn}$ maximum | dL | Δ | Tc |
|---|---|---|---|---|---|
| .05 | 4 | 15.2 | 26.7 | .9 | 550 |
| .5 | 3.9 | 13.3 | 26.6 | .064 | 560 |
| 1 | 4 | 15.6 | 27.9 | .074 | 550 |
| 1.5 | 4 | 13.9 | 26.1 | .077 | 510 |
| 2.0 | 4 | 9.94 | 35.1 | .07 | 530 |
| 2.5 | 4.8 | 7.61 | 32.22 | .071 | 590 |
| 3 | 4.7 | 5.97 | 28.77 | .066 | 570 |
| 3.5 | 4.9 | 6.39 | 25.9 | .073 | 510 |
| 4 | 5 | 7.19 | 24.8 | .07 | 490 |

Table-1: Maximum $d_{xn}$ versus $\epsilon$, L=1

In figure-15, the x & y trajectories of agent-1 resulting from the suggested protocol (single integrator) with N=200, L=1 and $\epsilon$=1 are shown. Despite the existence of few isolated sharp turns, the trajectories are well-behaved.

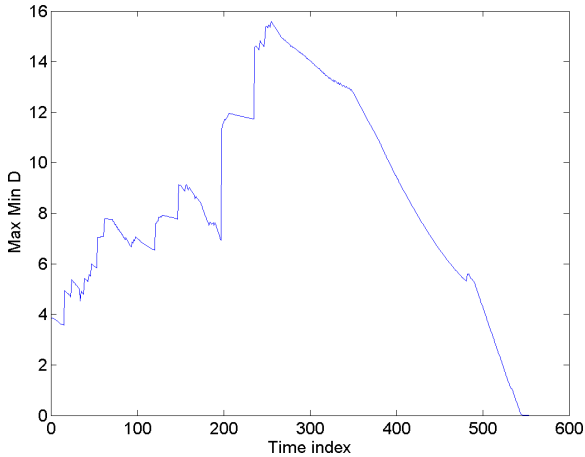
Figure-13: $d_{xn}$ versus time, $\epsilon$=1

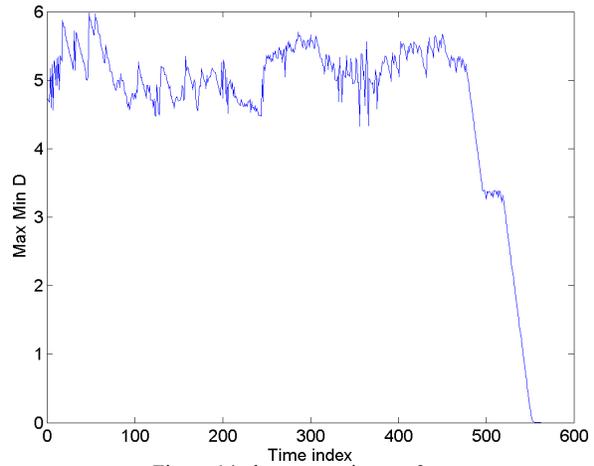
Figure-14: $d_{xn}$ versus time, $\epsilon$=3

| L | $d_{xn}$ initial | $d_{xn}$ maximum | dL | Δ | Tc |
|---|---|---|---|---|---|
| 1 | 4 | 15.6 | 27.6 | .0622 | 550 |
| 2 | 4 | 13.1 | 23.4 | 7.3 | 270 |
| 3 | 4 | 15.9 | 24.5 | 4.3 | 180 |
| 4 | 4 | 13.4 | 28.3 | 6.2 | 135 |
| 5 | 4 | 15.3 | 24.7 | 5.2 | 110 |
| 6 | 4 | 14.1 | 24.3 | 5.5 | 85 |
| 7 | 4 | 15.2 | 25.4 | 6.2 | 78 |
| 8 | 4 | 15 | 24.3 | 6.1 | 69 |
| 9 | 4 | 14.9 | 30.9 | 8 | 60 |
| 20 | 4 | 12.1 | 23.1 | 10.7 | 27 |
| 30 | 4 | 9.9 | 23.2 | 12.1 | 18 |
| 40 | 4 | 4.5 | 35.1 | 13.2 | 13 |

Table-2: Maximum $d_{xn}$ versus L, $\epsilon$=1.

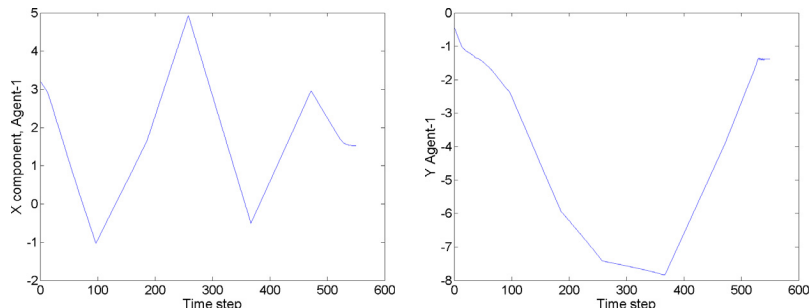
Figure-15, x & y trajectories of agent-1, suggested protocol.

It may be useful to examine the global behavior of the group that unfolds under the influence of the protocol. It is well-known that computing a closed form solution to determine *a priori* the exact global behavior of a complex dynamical system is not possible. However, the work of Christopher Langton [36] does provide useful insight regarding the effect of the parameter $\epsilon$ on global behavior. Langton observed that a small value of a protocol's parameter resulted in a chaotic group behavior while a high value of the parameter yielded an orderly group behavior. However, he found a value in-between where a small region lies on the border between order and chaos. This region generates a large number of metastable, complex, high-level structures. He called this region the edge of chaos. The $\epsilon$ parameter in the modified protocol is observed to have the same effect reported in [36]. The protocol is simulated for a large number of agents (2400 agents) creating a 4800 dimensional complex system. Three values of $\epsilon$ were used keeping the initial distribution of the agents the same for all cases. While in all cases the agents converged to the same position, the behavioral pattern each case exhibited towards consensus is different. For $\epsilon=0.25$ (figure-16) the behavior of the group during the consensus effort was totally chaotic. As for $\epsilon=2$ (figure-18), a predictable implosive action of the group was observed. Around $\epsilon=1$ (figure-17) the behavior of the group mutated yielding a large variety of complex structures before all agents converged to a single point.

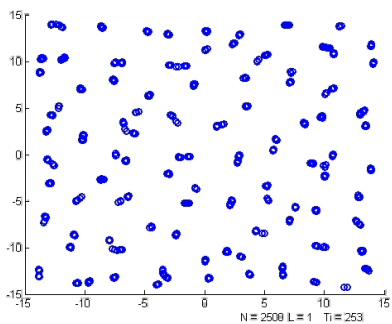 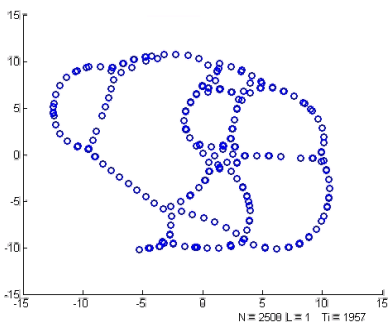 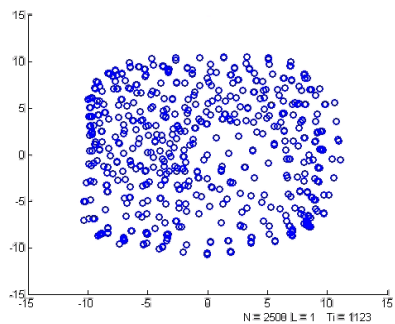

Figure-16: Random behavior, $\epsilon=.25$      Figure-17: Complex behavior, $\epsilon=1$      Figure-18: Deterministic behavior, $\epsilon=2$.

The ability of the techniques presented in section IV to convert the rendezvous guidance protocol into a rendezvous control protocol is tested. The directed communication graph in figure-19 is used for this purpose. The graph has a cycle that contains all the nodes. Therefore, convergence is guaranteed for a single integrator system.

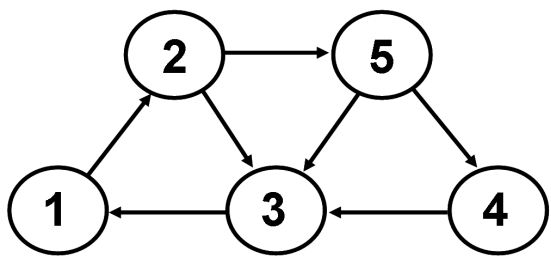

Figure-19: A directed communication graph

Figure-20 shows the response of five single integrator agents attempting to meet at the same location. As can be seen, the group converges to a rendezvous point. In figure-21, the single integrator agents were replaced by double integrator agents. As can be seen, the group failed to converge.

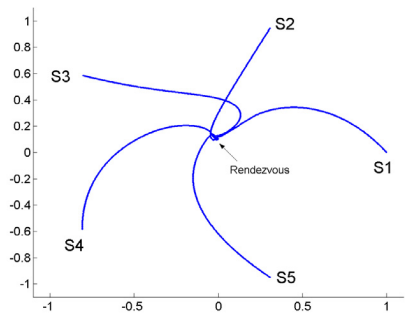

Figure-20: Single integrator agents using the graph in figure-19.

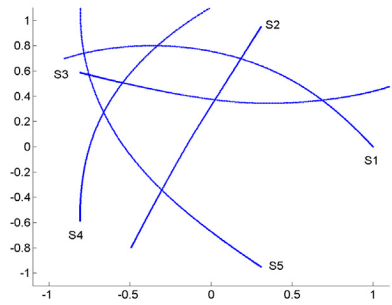

Figure-21: Double integrator agents using the communication graph in figure-19

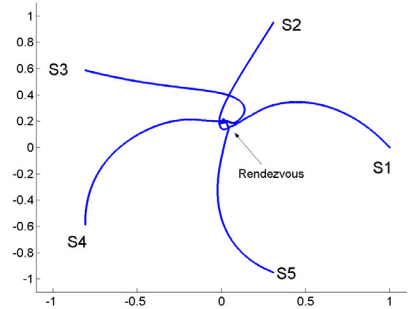

Figure-22: Double integrator agents using the communication graph in figure-19, NADF used.

In figure-22, the NADF approach is used to generate the rendezvous control protocol. The following parameters are selected $K_d=150$ and $b=2$. As can be seen, the resulting trajectories for the double integrator agents are almost identical to those of the single integrator agents. The control signals for the first agent are shown in figure-23. Despite the use of excessively high NADF, the control signal is well behaved. The convergence rate is also unaffected. The sharp fluctuations in the control signals that occur at the end are caused by interaction forces when the agents are in very close proximity to each other. The problem can be easily solved by requiring convergence to be to a small region instead of a point. Figure-24 shows the agents response when $K_d=0$ and $b=2$. As can be seen a high level of transients appeared.

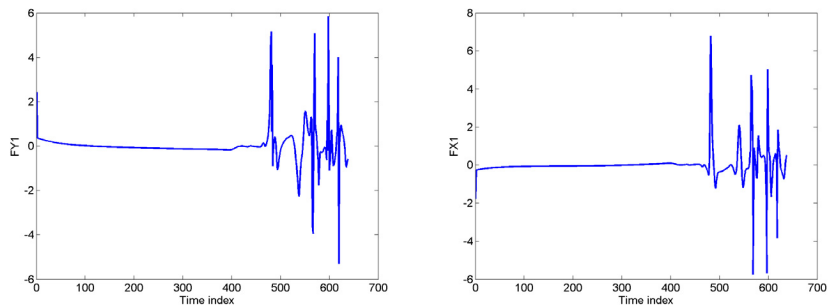

Figure-23: x & y components of the control signal for agent-1.

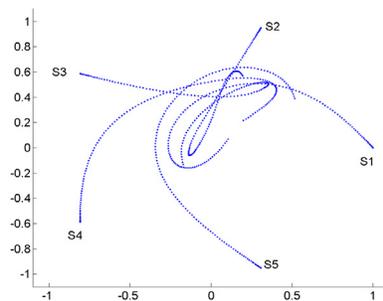

Figure-24: Double integrator agents using the communication graph in figure-, $K_d=0$, $b=2$.

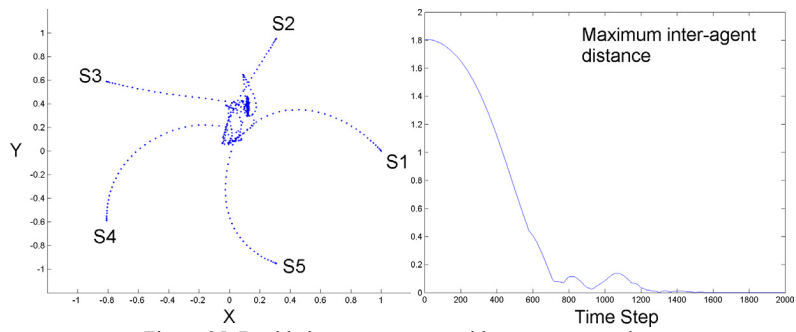
Figure-25: Double integrator agents with actuator saturation.

In figure-25 the effect of saturation on the protocol is tested. The same example in figure-22 is repeated with a saturation nonlinearity limiting the components of the control signal to a maximum absolute value of 1 for all agents. The maximum forces for the unsaturated case are fx=15.5 and fy =12.5 (more than 80% saturation). As can be seen from the agents' trajectories as well as the maximum inter-agent distance profile this high level of saturation has minimum effect on the ability of the agents to reach a common meeting point.

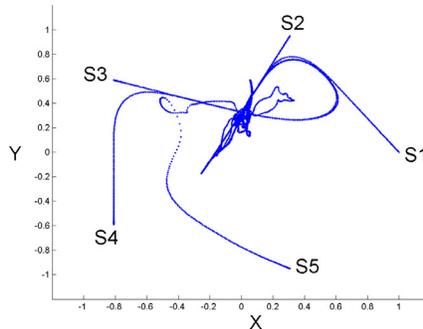
Figure-26: double integrator agents with 2 seconds communication delay

The effect of a delay in communication is also tested in figure-26. The example in figure-22 is repeated with a delay of 2 seconds. As can be seen, the agents were still able to reach a meeting point and the trajectories are relatively well-behaved. In figure-27 the maximum inter-agent distance is plotted as a function of time for different communication delays. As can be seen, the agents were able to reach consensus in all cases. The only effect of the delay seem to be a longer convergence time.

The effect of external drift forces on the rendezvous control protocol is tested. The example in figure-22 is repeated with the following drift forces applied to the agents. ($fdx_1$=0, $fdy_1$=1), ($fdx_2$=1, $fdy_2$=1), ($fdx_3$=-1, $fdy_3$=1), ($fdx_4$=-1, $fdy_4$=0), ($fdx_5$=0, $fdy_5$=-1). As can be seen from figure-28, the agents still managed to reach consensus and the trajectory remained well-behaved. In figure-29 the drift forces were augmented with a 2 second delay in communication. The ability of the group to reach a rendezvous location is still not affected.

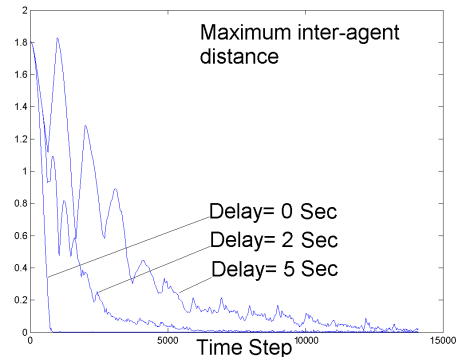
Figure-27: Maximum inter-agent distance versus time for different communication delays, double integrator agents.

Although the protocol was not designed for cooperative tracking, it does demonstrate a promising potential in this regard. The same example in figure-22 is used. The motion pattern to be tracked is injected into the group by directly controlling the motion of the fifth agent. Figure-30 shows the remaining four agents closely tracking a linear and a sinusoidal motion patterns.

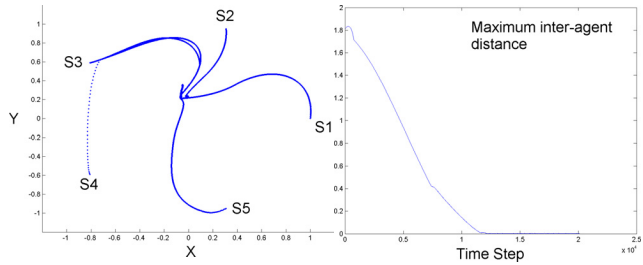
Figure-28: Double integrator agents with external drift

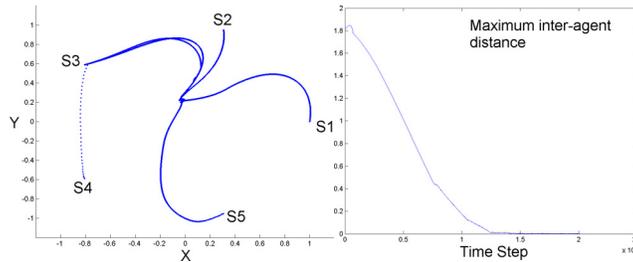
Figure-29: Double integrator agents subject to both external drift and 2 seconds communication delays

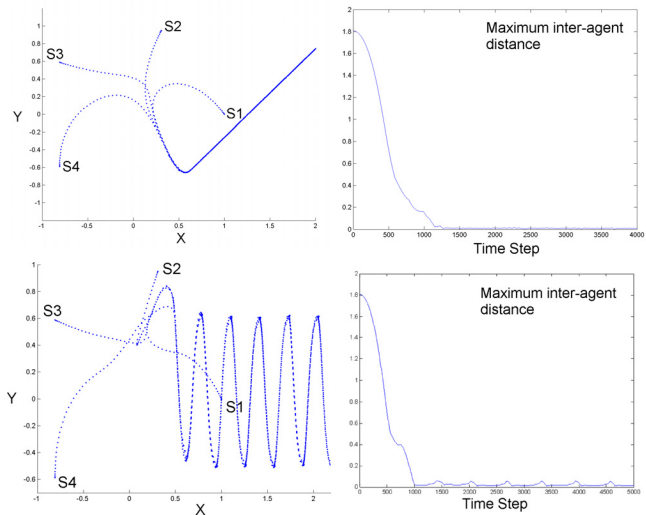
Figure-30: Cooperative tracking, double integrator agents.

The reason a fixed communication graph is used in the previous examples is to demonstrate that the control protocol will guarantee convergence as long as the rendezvous guidance protocol for a single integrator system is convergent. All reported results are repeatable for the random communicating graph generated by the suggested protocol. Figure-31 shows the previous configuration of the five single integrator agents reaching a rendezvous point using the suggested protocol.

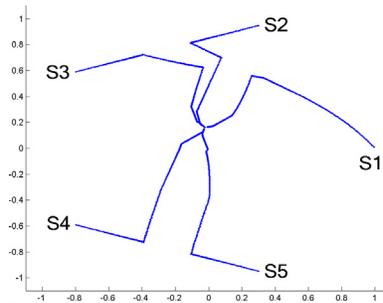
Figure-31: five single integrator agents using the suggested protocol

In figure-32 the agents are replaced with double integrator agents. The agents generate the control signal using the suggested control for the same the previous choice of parameters. As can be seen, the trajectories the agents took to the rendezvous point are almost identical to the single integrator case in figure-31.

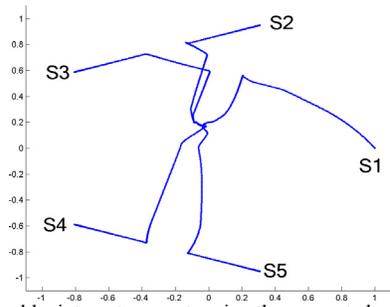
Figure-32: five double integrator agents using the suggested control protocol

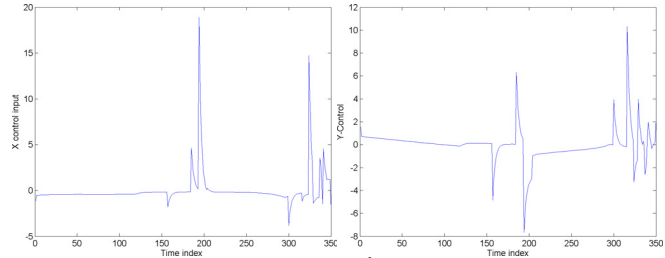
Figure-33: Control signals for the 5$^{th}$ agent in figure-32.

In figure-34 80% actuator saturation is added. As can be seen the trajectories did not get significantly affected. The corresponding control signals are shown in figure-35.

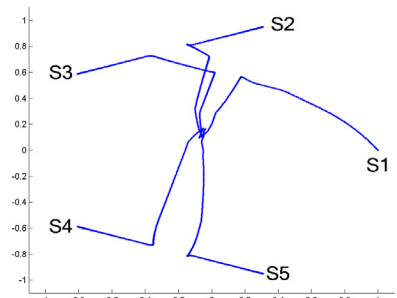
Figure-34: agents in figure-32 with 80% actuator saturation.

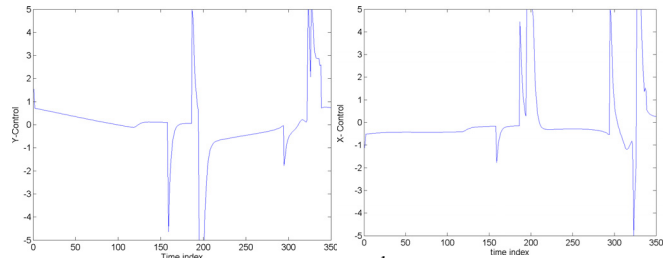
Figure-35: Control signals for the 5$^{th}$ agent in figure-34.

The ability of the control scheme suggested in equation-50 [31] to convert the guidance signal into a rendezvous control signal for a group of car-like robots is tested in figure-36. A group of five agents that are uniformly distributed on the circumference of a circle with unity radius, all initially oriented along the positive x-axis are used. The agents are using the communication graph in figure-19 to exchange data. The parameters used are $K_1$=0.5 and $K_2$=4. The orientation and control signals of the fifth agent are shown in figures-37,38 respectively. Figure-39 shows the trajectories of the agents for a different set of parameters $K_1$=0.5, $K_2$=10. $K_2$ is responsible for improving the alignment of the robot with the guidance field. Increasing it does improve the response of the group.

Figure-40 shows two jets described by the system equation in (29) taking-off and synchronizing their orientation in a decentralized manner by converting the guidance signal into a control signal using the method suggested in [32]. The control signals for jet-2, banking angle, tangential force and normal force are shown in figures 41,42,43 respectively.

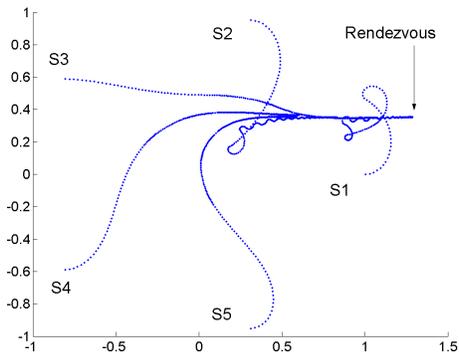

Figure-36: trajectories for car-like agents, $K_1=.5$ $K_2=4$

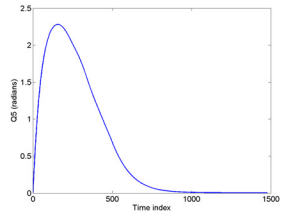 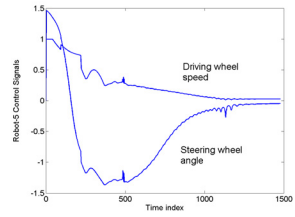

Figure-37: Orientation, agent-5    Figure-38: Control signals agent-5

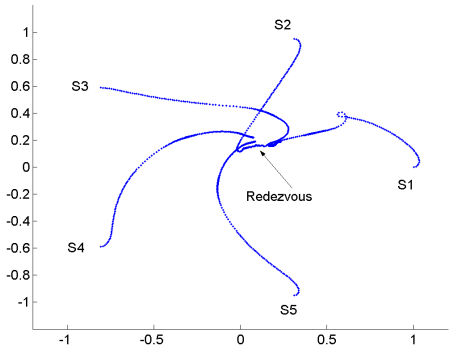

Figure-39: trajectories for car-like agents, $K_1=.5$, $K_2=10$

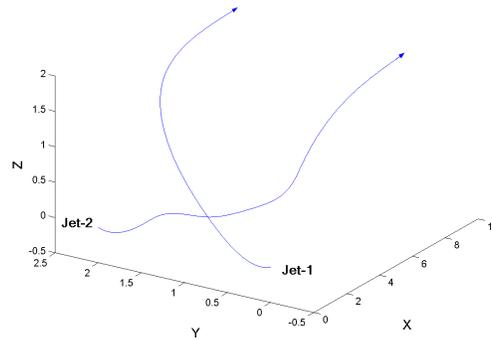

Figure-40: two jets synchronizing their orientations.

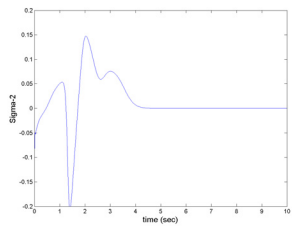 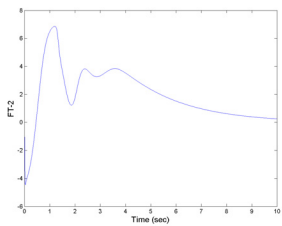 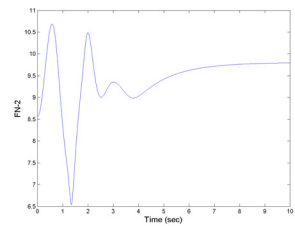

Figure-41, banking angle, Jet-2    Figure-42, tangential force, jet-2    Figure-43, normal lift force, jet-2

# VI. Conclusions

A new nearest neighbor, opportunistic, control protocol is suggested to enable a group of mobile agents to reach a common location in an RF-challenged and GPS-denied environment where assumptions about future connectivity are not possible. The protocol is practical and can guarantee convergence under sparse communication conditions. It also have the ability to prevent the growth of the communication burden during the effort to reach a common meeting point. It is shown that the guidance signal from the protocol, which suits single integrator dynamics, may be converted into a provably-correct control signal for involved dynamical agents such as second order dynamical systems, UGVs and UAVs. Each agent does the conversion from guidance to control using its velocity only. This is important since the velocities of other agents are difficult, if at all possible, to estimate when connectivity is random. The control protocol is demonstrated to be robust to communication delays and actuator saturation . It also produces good results when external drift is present.

**Acknowledgment:** The author would like to thank king Fahd university of petroleum and minerals for its support.